\documentclass[10pt,journal,compsoc,two column]{IEEEtran}%
\usepackage{amsfonts}
\usepackage{amsmath}
\usepackage{graphicx}
\usepackage{subfigure}
\usepackage{multirow}
\usepackage{amssymb}
\usepackage{epstopdf}
\usepackage{placeins}
\usepackage{booktabs}
\usepackage{algorithmic}
\usepackage{placeins}
\usepackage{lineno}
\usepackage{hyperref}
\usepackage[nocompress]{cite}
\usepackage{cite}%
\providecommand{\U}[1]{\protect\rule{.1in}{.1in}}
\ifCLASSOPTIONcompsoc
\else
\fi
\makeatother
\begin{document}

\title{A Unified Approach to Kinship Verification}
\author{~Eran~Dahan and Yosi~Keller~
\IEEEcompsocitemizethanks{\IEEEcompsocthanksitem  E. Dahan \& Y. Keller are with the Faculty of Engineering, Bar-Ilan University,
E-mail:yosi.keller@gmail.com}}
\maketitle

\begin{abstract}
In this work, we propose a deep learning-based approach for kin verification
using a unified multi-task learning scheme where all kinship classes are
jointly learned. This allows us to better utilize small training sets that are
typical of kin verification. We introduce a novel approach for fusing the
embeddings of kin images, to avoid overfitting, which is a common issue in
training such networks. An adaptive sampling scheme is derived for the
training set images to resolve the inherent imbalance in kin verification
datasets. A thorough ablation study exemplifies the effectivity of our
approach, which is experimentally shown to outperform contemporary
state-of-the-art kin verification results when applied to the Families In the
Wild, FG2018, and FG2020 datasets.

\end{abstract}








\IEEEpeerreviewmaketitle


\section{Introduction}

\label{sec:Introduction}

\begin{figure}[t]
\begin{center}
\centering\includegraphics[width=1.0\linewidth]{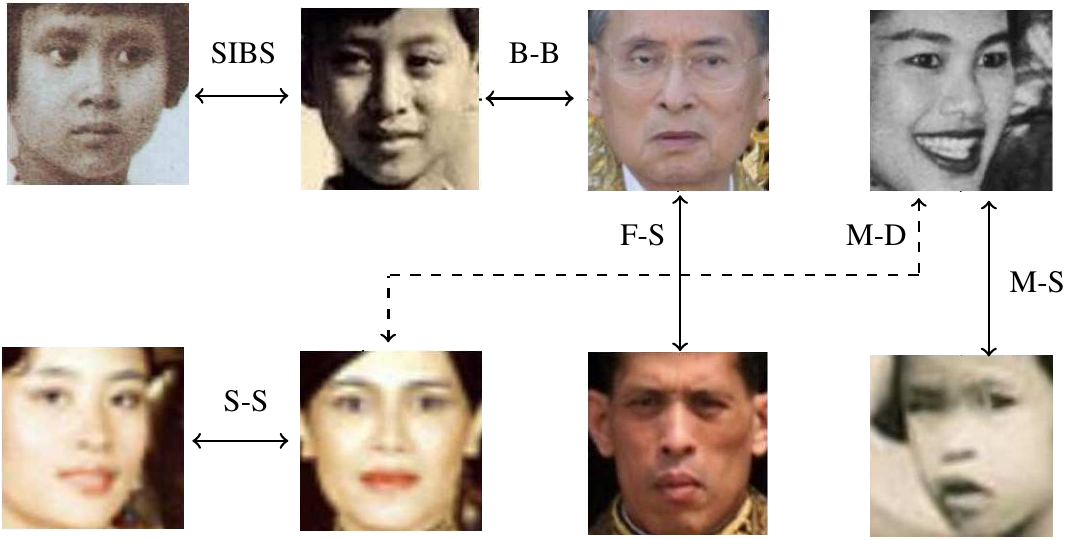}
\end{center}
\caption{The kinship verification task is to determine whether two face images
are related by a particular kinship class, such as Brothers (B-B), Sisters
(S-S), Father-Daughter (F-D), Father-Son (F-S), Mother-Son (M-S),
Mother-Daughter (M-D), and Siblings (SIBS).}%
\label{fig:main_idea}%
\end{figure}

The goal of kin verification \cite{familyclas,familyclas2,kinsim,kinsim2} is
to verify whether or not two people are related by a particular kin
relationship given their face images. The kinship classes are depicted in Fig.
\ref{fig:main_idea}, where given a pair of face images $\left\{  \mathbf{\phi
}_{i},\mathbf{\varphi}_{i}\right\}  $ and a kinship class $\lambda_{i}%
\in\Lambda=$\{B-B, F-S, M-D, S-S, F-D, M-S, SIBS\}{, }we aim to verify whether
$\mathbf{\phi}_{i}$ and $\mathbf{\varphi}_{i}$ are related by $\lambda_{i}$.
In contrast to face recognition, kin faces are non-identical, and their visual
similarity is often unintuitive, even for human observers, as kins might
differ by gender and notable age differences. The facial similarity\ of kins
varies significantly between different families and even within the kins of a
particular family, implying significant intraclass variability. As face
recognition schemes aim to identify individual subjects, kins will be
identified as different individuals. Following the common approach in face recognition, classical kin
verification schemes encoded the input faces $\left\{  \mathbf{\phi}%
_{i},\mathbf{\varphi}_{i}\right\}  $ using handcrafted features $\left\{
\widehat{\mathbf{\phi}}_{i},\mathbf{\widehat{\varphi}_{i}}\right\}  $
\cite{fang2010towards,Hand_SIFT_flow_based_genetic2016,Dandekar} such as LBP,
HOG, and Gabor. The representation was refined using metric-learning
\cite{Liris-7042,Haibin,Metric_neighborhood_repulsed2017}, and the kinship was
classified by applying KNN and SVM classifiers. With the emergence of deep
learning, it was applied to kin verification. Facial features were derived by
training face recognition CNNs on large-scale face recognition datasets
\cite{DEEP_From_face_recognition_to_kinship_verification2017} and using
transfer learning to apply them to kin verification
\cite{DEEP_From_face_recognition_to_kinship_verification2017}. Recent results
by Wang et al. \cite{SMDL_ref} applied a Generative Adversarial Net (GAN) to
mitigate the age-differences by synthesizing face images depicting
intermediate ages. Small-scale datasets were initially utilized in kin
verification \cite{XIA,FG2015Kinship,fang2010towards}, most of which were
shown \cite{7393855,Dawson_2019} to be biased, as the faces of kins were often
cropped from the same photo, implying that the kinship could be inferred from
the similar chromatic properties of the photo crops. The large-scale RFIW
dataset \cite{fiwpamiSI2018} was the only one shown to be unbiased, consisting
of kin pairs cropped from different photos. The novel FIW in Multimedia
(FIW-MM) database, recently introduced by Robinson et al.
\cite{Robinson2020FamiliesIW}, is a large-scale multi-modal kin verification
dataset, consisting of of video, audio, and contextual transcripts

In this work, we propose a deep learning-based approach for kin verification
using a siamese CNN. We apply multi-task learning to train a single unified
CNN for \textit{all} kinship classes, allowing to leverage their training
samples to refine the face embedding backbone CNN. We show that kin
verification, in contrast to face recognition, is prone to overfitting and
derive a novel CNN architecture for the fusion of face embeddings based on
transposed $1\times1$ convolutions. We also show that kin verification
datasets are inherently imbalanced in terms of the number of samples per
family, and samples per kin (father, mother, etc.), implying that the training
process might be biased. For that, we propose an adaptive sampling scheme for
the training set, which is shown to improve the kin verification accuracy.
Some kin verification classes, such as B-B and D-D, are symmetrical, allowing
to utilize siamese CNNs. In contrast, other kinship classes, such as F-D, F-S,
etc., are asymmetric in terms of age and gender, requiring different CNNs for
each class. For that, we applied an asymmetric weighting layer that improves accuracy.

In particular, we propose the following contributions: (1) We present a
unified approach for kin verification that applies multi-task learning to
jointly utilize the training sets of all kinship classes. (2) We show that kin
verification is prone to overfitting and suggest a novel subnetwork for the
fusion of feature maps based on cascaded $1\times1$ convolutions. (3) To
overcome the imbalanced training set, we introduce an adaptive approach for
sampling the training pairs, that is shown to improve the kin verification
accuracy. (4) The proposed scheme is experimentally shown to outperform
contemporary state-of-the-art approaches, when applied to the RFIW
\cite{fiwpamiSI2018}, FG2018, and FG2020 \cite{robinson2020recognizing} datasets.

\section{Related Work}

\label{sec:RELATED}

Kin verification is related to face verification, and similar approaches were
often used \cite{fiwpamiSI2018}. Thus, early kin verification schemes utilized
handcrafted image descriptors for encoding the face images. Puthenputhussery
et al\textit{.} \cite{Hand_SIFT_flow_based_genetic2016} used SIFT descriptors
and genetic fisher vectors to optimize the similarity between true and false
kins. Similarly, Fang et al. \cite{fang2010towards} applied LBP, HOG, and
Gabor descriptors to encode the face images and trained KNN and Kernel SVM
classifiers. LBP was used as a local face descriptor by Dander and Limbate,
\cite{Dandekar}. The informative patches in the face images were detected by
Quin et al. \cite{Qin} by applying sparse regularized group lasso regression
to a SIFT-based face descriptor. The resulting face embeddings were classified
by a linear SVM.

Mahpod and Keller
\cite{Metric_multiview_hybrid_distance2018,Dahan2017KinVerificationMO}
proposed a metric-learning-based approach, applying symmetric and asymmetric
metric learning to face embeddings using handcrafted features. The resulting
representations were classified by an SVM classifier, and the scheme was
applied to the KinFaceW-I and KinFaceW-II datasets \cite{FG2015Kinship}. A
color histogram was found to be the most informative image feature,
corresponding to the later findings by Lopez et al. \cite{7393855} and Damson
\textit{et al.} \cite{Dawson_2019} who showed these datasets, among others, to
be color-biased. A neighborhood repulsed correlation-metric-learning approach
was used by Yang \cite{Metric_neighborhood_repulsed2017}, while Xn and Shan
proposed an ensemble of bilinear models
\cite{Metric_structured_similarity_fusion2016}, where each model was trained
by metric-learning using multiple face descriptors. Handcrafted image
descriptors, such as LBP and Fisher vectors, were used in the Triangular
Similarity Metric-Learning (TSML) approach by Pilei et al. \cite{Liris-7042},
where the dimensions of the descriptors were reduced by metric learning.
Optimal weights per feature were computed by Hairpin et al. \cite{Haibin} in
the Discriminative Multimetric-Learning (DML) scheme, and the weighted
features were used within a metric learning formulation.

With the emergence of deep learning, CNNs were also applied to kin
verification. Lu et al. \cite{Deep_Discriminative_deep_metric_learning2017},
proposed discriminative deep metric-learning (DML) to train a CNN to jointly
learn multiple metric-learning networks, and use ensemble-learning to optimize
the results. An asymmetric scheme, where each image is processed by a
different CNN that is adjusted to the input (Father, Son, etc.), was proposed
by Avignon et al. \cite{CMML}, where metric-learning was applied to the face
embeddings. A coarse-to-fine domain adaptation approach was derived by Duad et
al. \cite{DEEP_From_face_recognition_to_kinship_verification2017} to utilize
the deep features learned by a face recognition network as kin verification features.

A large-scale kin verification evaluation by human reviewers was conducted by
Kohl et al. \cite{Deep_Hierarchical_representation_kinship_verification2017},
who proposed a hierarchical verification scheme, using representation learning
to encode the face regions. The Kinnet CNN was introduced by Li et al.
\cite{kinnet_ref}, where the face embeddings were learned using a large-scale
face recognition dataset. A triplet loss was applied to learn the kin
verification, and the number of images per family member was balanced by image
augmentation. An ensemble of four CNNs was applied to improve the accuracy.
Kohl et al. \cite{Deep_Supervised_mixed_norm_autoencoder2017} applied
autoencoders to detect kin similarity in unconstrained videos, using a
supervised mixed-norm autoencoder to compute the sparse embeddings of the kin
pairs. Deep autoencoders were also studied by Deadman et al. \cite{Wang:2016}
to detect informative facial features for kin verification, where a Sparse
Discriminative Metric Loss (SDM-Loss) was derived to utilize the positive and
negative training pairs. A review and evaluation of contemporary schemes were
detailed in the RFIW 2020 challenge \cite{robinson2020recognizing}, organized
by Robinson et al..

A Generative Adversarial Net (GAN) was proposed by Wang et al. \cite{SMDL_ref}
to mitigate the age differences by synthesizing younger and older face images
of parents and their siblings, respectively. Thus, a conditional GAN\ was used
to synthesize cross-generation kins in an intermediate age domain, to improve
the face similarity. Ozkan and Ozkan \cite{Kinshipgan} proposed a GAN-based
approach for synthesizing an image of a child's face using the face images of
a single parent. The synthesis is derived by a generator network in an
auto-encoder architecture. An adversarial loss was applied using large-scale
unsupervised data, to mitigate the overfitting, due to the small annotated
training sets. The stability of the approach was improved using
cycle-consistency. Gao et al. \cite{gao2019child} introduced a novel
biologically-motivated kinship generation approach, where the facial
appearance of a child is inherited from both parents by a random genomic
fusion process. For that, a three-step approach was derived, where first a
conditional auto-encoder (CAD) is trained on a large-scale face dataset to
encode the facial appearance given age and gender. Second, a CNN\ denoted as
DNA-Net is trained to further encode the face embeddings into an embedding
denoted as genes. Random subsets of male and female genes are combined and
decoded by the CAD decoder to synthesize the child's face.

A fundamental flaw in most contemporary kin and family verification image
datasets was detected by Lopez et al. \cite{7393855}, who noticed that the
KinFaceW datasets were created by cropping the kin face images from the same
family photographs. Hence, learning-based schemes in general, deep
learning-based in particular, might learn the chromatic (\textquotedblleft
from same photo\textquotedblright) cues instead of the kinship similarity.
Dawson et al. \cite{Dawson_2019} extended these results by training the deep
learning-based From Same Photo (FSP) image classifier that was trained to
detect whether two image patches were cropped from the same photo. The
FSP\ was applied to most contemporary kin face datasets, achieving accurate
kinship classification in all, but for the RFIW dataset. Thus, in this work,
we evaluate the proposed scheme by training it, using only the RFIW dataset
\cite{fiwpamiSI2018}. A thorough survey by Robinson et al.
\cite{Robinson2020VisualKR} details recent approaches and datasets in kin
verifcation.\begin{figure}[t]
\centering\includegraphics[width=1.0\linewidth]{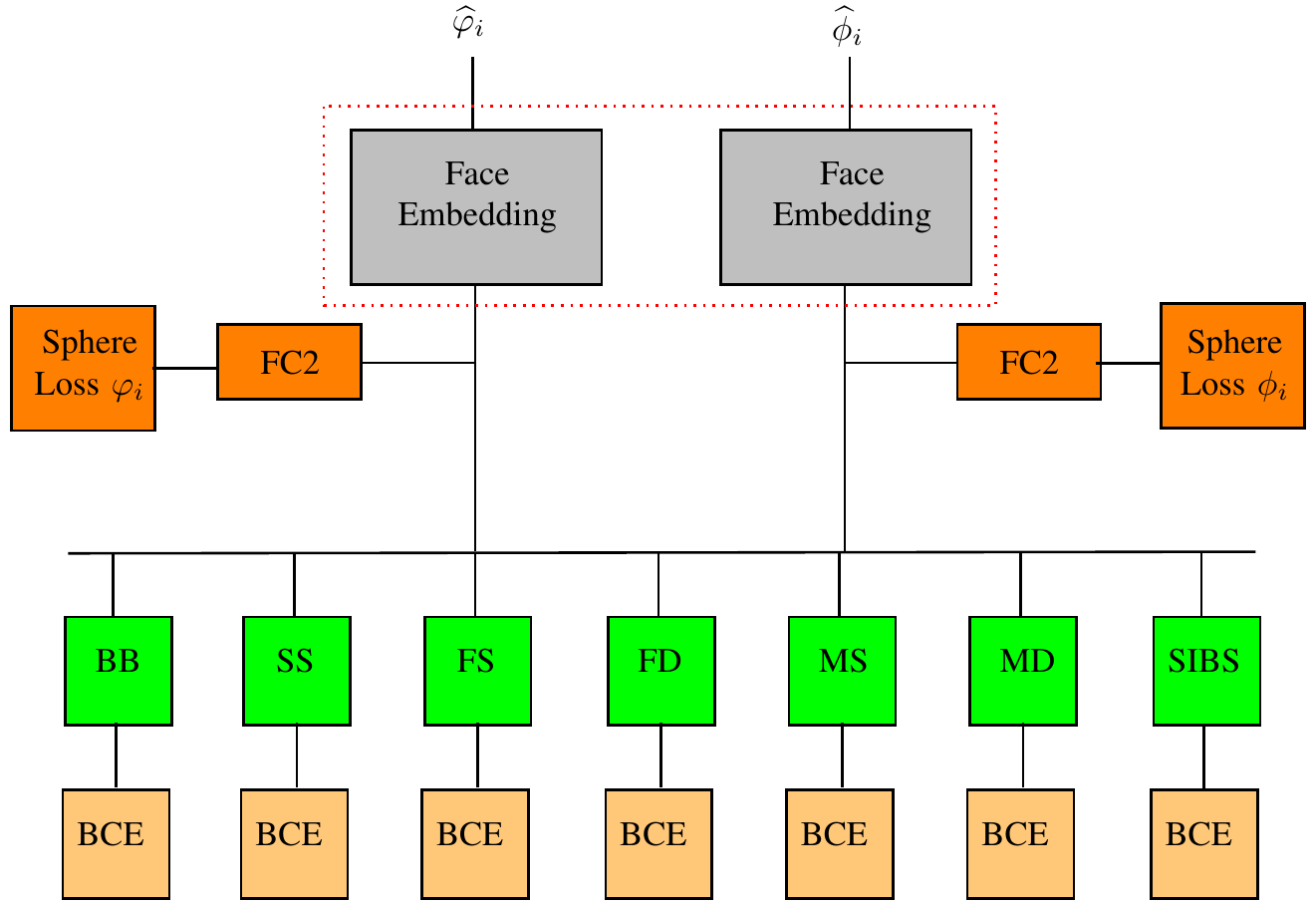}\caption{The
proposed kin verification system. The logical components are color-coded. The
face embedding CNN is shared by all kinship classes and is initially
pretrained on the CASIA-WebFace face image dataset \cite{CASIA}. The
embeddings $\mathbf{\widehat{\varphi}}_{i}$ and $\widehat{\mathbf{\phi}}_{i}$
of the input face images are fused and classified by separate subnetworks, one
per kinship class. The weight-sharing FC-2 layer is used to refine the face
embedding on the RFIW dataset \cite{fiwpamiSI2018}.}%
\label{fig:total_arc}%
\end{figure}

\section{Unified Multi-Task Kin Verification}

\label{sec:unified-mt}

In this work, we propose a unified multi-task kin verification approach, where
all kinship classes are jointly learned, using the CNN\ depicted in Fig.
\ref{fig:total_arc}. It consists of a siamese network for learning the face
embeddings used in the classification of all of the kinship classes, followed
by multiple kinship classification subnetworks, one per kin verification task,
e.g., B-B, SIBS.

The embeddings $\left\{  \mathbf{\widehat{\varphi}}_{i},\widehat{\mathbf{\phi
}}_{i}\right\}  $ of $\left\{  \mathbf{\phi}_{i},\mathbf{\varphi}_{i}\right\}
$ are computed using the siamese CNN detailed in Section
\ref{subsec:embedding}. $\left\{  \mathbf{\widehat{\varphi}}_{i}%
,\widehat{\mathbf{\phi}}_{i}\right\}  $ are jointly trained end-to-end using
\textit{all} kinship classes and corresponding images in the training set. In
contrast, in previous schemes \cite{fiwpamiSI2018}, a separate classification
CNN was trained per kinship class, utilizing an order of magnitude less
training images. Each kin verification triplet $\left\{  \mathbf{\widehat
{\varphi}}_{i},\widehat{\mathbf{\phi}}_{i},\lambda_{i}\right\}  $ is
classified by the embedding-fusion and classification subnetwork depicted in
Fig. \ref{fig:total_arc}. There are $\left\vert \Lambda\right\vert =7$ such
subnetworks that are detailed in Section \ref{subsec:siamese_asymmetric}. A
novel embedding-fusion approach is introduced in Section
\ref{subsec:siamese_asymmetric}, based on channel-wise $1\times1$
convolutions, to avoid overfitting. The overfitting issue was previously
reported by Robinson et al. \cite{fiwpamiSI2018} when applying distance
learning to face embeddings computed using the Sphereface CNN, resulting in
little classification accuracy improvement. Kin verification datasets in
general, and the RFIW dataset \cite{fiwpamiSI2018} in particular, are
inherently imbalanced, with respect to the number of images per family, and
the number of images per kin, in each family. For that, we propose in Section
\ref{subsec:adaptive-sampling} an adaptive image sampling scheme that is shown
to improve the verification accuracy.

The proposed scheme is trained using multiple losses. The siamese face
embedding CNN is refined using the Sphere losses ${L}_{Sphere}^{\widehat
{\mathbf{\phi}}}$ and ${{L}_{Sphere}^{\widehat{\varphi_{i}}},}$ that are
applied to classify the embeddings $\left\{  \mathbf{\widehat{\varphi}}%
_{i},\widehat{\mathbf{\phi}}_{i}\right\}  $ of the 10,676 identities in the
RFIW dataset. The kin verification is classified using the $\left\vert
\Lambda\right\vert =7$ subnetworks trained using $\left\{  {{L}_{BCE}%
^{\lambda_{i}}}\right\}  _{\lambda_{i}\in\Lambda}$ Binary Cross-Entropy (BCE)
losses, such that ${{L}_{BCE}^{\lambda_{i}}}$ verifies the kinship
$\lambda_{i}$. Given a particular training triplet $\left\{  \mathbf{\widehat
{\varphi}}_{i},\widehat{\mathbf{\phi}}_{i},\lambda_{i}\right\}  $, all other
BCE losses are zeroed%
\begin{equation}
\left\{  {{L}_{BCE}^{\lambda_{k}}}\right\}  =0,\text{ }\forall\lambda_{k}%
\in\Lambda,\lambda_{k}\neq\lambda_{i}.
\end{equation}
Thus, the overall loss is given by
\begin{multline}
L\left(  \mathbf{\widehat{\varphi}}_{i},\widehat{\mathbf{\phi}}_{i}%
,\lambda_{i}\right)  =\label{equ:overall loss}\\
{L}_{Sphere}^{\widehat{\mathbf{\phi}}}+{{L}_{Sphere}^{\widehat{\varphi_{i}}}%
}+\alpha^{2}%
{\displaystyle\sum\limits_{\lambda_{k}}}
{{L}_{BCE}^{\lambda_{k}}}\cdot{\Delta}\left(  \lambda_{k}-\lambda_{i}\right)
\end{multline}
where%
\begin{equation}
{\Delta}\left(  x\right)  =\left\{
\begin{tabular}
[c]{ll}%
$1,$ & $x=0$\\
$0,$ & $else$%
\end{tabular}
\ \ \right.  ,\text{ }x\in%
\mathbb{R}
.
\end{equation}
and $\alpha\in%
\mathbb{R}
$. We tested a range of values of $\alpha$, and the kin verification accuracy
was robust to its choice. Hence, $\alpha=1$ was used.

\subsection{Face Embedding}

\label{subsec:embedding}

The backbone of the face embedding CNN is based on the Sphereface CNN
\cite{Liu_2017}, which was used by Robinson et al. \cite{fiwpamiSI2018} to
achieve state-of-the-art kin-verification results. It is detailed in Table
\ref{tab:BlockTable} where the embedding of an input image $\mathbf{\varphi}$
is given by the output of the FC-1 layer $\widehat{\mathbf{\varphi}%
}\mathbf{\in}%
\mathbb{R}
^{512}$. The output of FC-2 is used to refine the face embedding CNN using the
RFIW dataset \cite{fiwpamiSI2018} and is optimized by the {SphereLoss
}\cite{Liu_2017}. Each training sample consists of a pair $\left\{
\mathbf{\phi}_{i},\mathbf{\varphi}_{i}\right\}  $ of input images, and both
CNNs are jointly trained via weight sharing.

\begin{table}[tbh]
\caption{The face embedding network. The CNN is based on the Sphereface CNN
\cite{Liu_2017}. Residual units are shown in double-column brackets, and S2
denotes stride 2. The output of FC-1 is used as the embedding $\widehat
{\mathbf{\varphi}}$ of the input image $\mathbf{\varphi}$. The output of the
layer FC-2 is used to refine the face embedding CNN. }%
\label{tab:BlockTable}
\begin{center}
{\small \
\begin{tabular}
[c]{|c|c|}\hline
Layer & 20-layer Sphereface CNN\\\hline\hline
\multicolumn{1}{|l|}{Conv1.x} & $\,\,%
\begin{tabular}
[c]{l}%
$\ [3\times3,64]\times1,S2$\\
$%
\begin{bmatrix}
3\times3,64\\
3\times3,64
\end{bmatrix}
\,\,\times1\,\,\,$%
\end{tabular}
\,$\\\hline
\multicolumn{1}{|l|}{Conv2.x} &
\begin{tabular}
[c]{l}%
$\ \ \ [3\times3,128]\times1,S2$\\
$%
\begin{bmatrix}
3\times3,128\\
3\times3,128
\end{bmatrix}
\,\,\times2\,\,\,$%
\end{tabular}
\\\hline
\multicolumn{1}{|l|}{Conv3.x} &
\begin{tabular}
[c]{l}%
$\ \ [3\times3,256]\times1,S2$\\
$%
\begin{bmatrix}
3\times3,256\\
3\times3,256
\end{bmatrix}
\,\,\times4\,\,\,$%
\end{tabular}
\\\hline
\multicolumn{1}{|l|}{Conv4.x} &
\begin{tabular}
[c]{l}%
$\ \ [3\times3,512]\times1,S2$\\
$%
\begin{bmatrix}
3\times3,512\\
3\times3,512
\end{bmatrix}
\,\,\times1\,\,\,$%
\end{tabular}
$\,\,\,$\\\hline
\multicolumn{1}{|l|}{FC-1} & $512$\\\hline
\multicolumn{1}{|l|}{FC-2} & $10,676$\\\hline
\multicolumn{1}{|l|}{SM} & $10,676$\\\hline
\multicolumn{1}{|l|}{Sphere loss} & $10,676$\\\hline
\end{tabular}
}
\end{center}
\end{table}

\subsection{The Fusion of Face Embeddings}

\label{subsec:siamese_asymmetric}

\begin{figure}[tbh]
\centering%
\begin{tabular}
[c]{l}%
\subfigure[]{\includegraphics[width=1.0\linewidth]{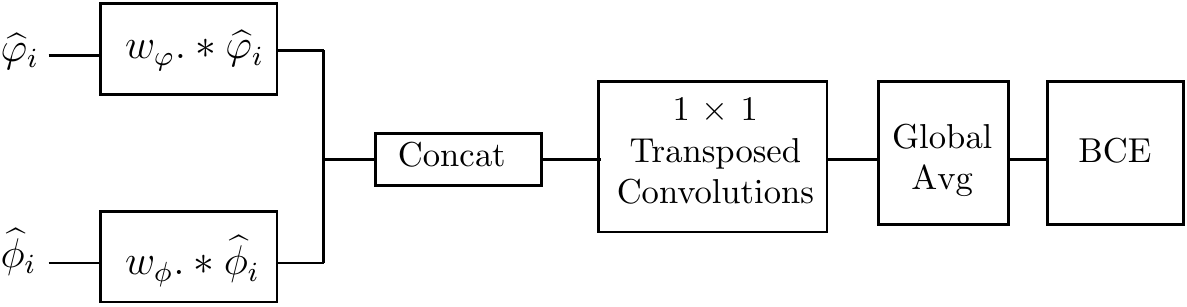}}\\
\subfigure[]{\includegraphics[width=1\linewidth]{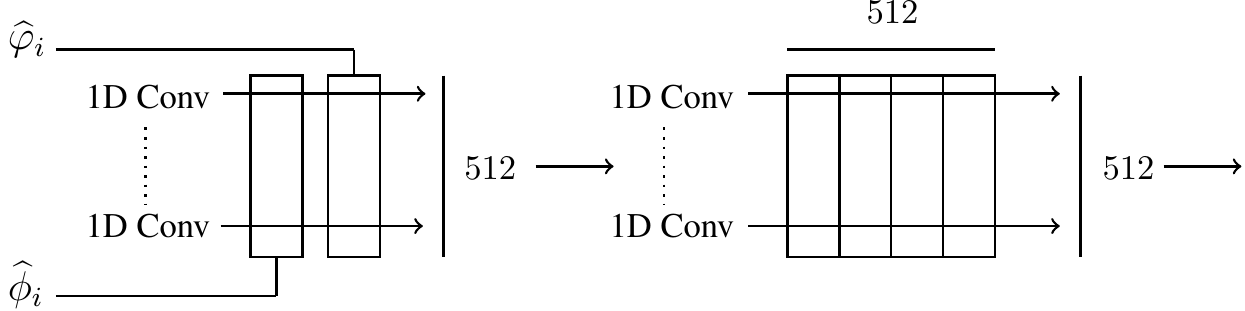}}
\end{tabular}
\caption{The kinship classification subnetwork. Multiple such subnetworks are
applied, one per kinship class. (a) The face embeddings $\mathbf{\widehat
{\varphi}}_{i}$ and $\widehat{\mathbf{\phi}}_{i}$, are pointwise weighted
($.\ast)$ by the learnt weights $w_{\varphi}$ and $w_{\mathbf{\phi}}$,
respectively. The weighted embeddings are processed by the fusion subnetwork
in Table \ref{tab:residual_conv}. Average pooling and a Softmax activation are
applied to the fused embedding, before the BCE loss. (b) The input layers of
the proposed embedding fusion subnetwork. The embeddings $\mathbf{\widehat
{\varphi}}_{i}\mathbf{\in}\mathbb{R}^{512}$ and $\widehat{\mathbf{\phi}}%
_{i}\mathbf{\in}\mathbb{R}^{512}$ are concatenated channel-wise, resulting in
a $\mathbf{w\in}\mathbb{R}^{512\times2}$ activation map, that is processed by
512 $1\times1$ transposed (channel-wise) convolutions, resulting in an
\thinspace$\mathbb{R}^{512\times512}$ activation$.$}%
\label{fig:inside_model}%
\end{figure}

The face embeddings $\left\{  \mathbf{\widehat{\varphi}}_{i},\widehat
{\mathbf{\phi}}_{i}\right\}  \mathbf{\in}%
\mathbb{R}
^{512}$ are fused to compute the kin verification scores. For that, we propose
a novel fusion subnetwork, whose core attribute is avoiding overfit that is
common in kin verification CNNs due to the small training sets, and
significant intraclass variability. An overview of the proposed fusion
subnetwork is shown in Fig. \ref{fig:inside_model}. Concatenation and bilinear pooling \cite{lin2015bilinear} are
common embedding fusion approaches that are followed by corresponding FC
layers. This requires a significant number of training parameters, resulting
in overfitting, as exemplified in Section \ref{subsec:Ablation}. Hence, we
propose a novel low-capacity fusion subnetwork, that utilizes significantly
fewer parameters. It is shown to provide better generalization and improves
the verification accuracy. The core of the proposed fusion approach is
depicted in Fig. \ref{fig:inside_model}b, where the inputs $\left\{
\mathbf{\widehat{\varphi}}_{i},\widehat{\mathbf{\phi}}_{i}\right\}
\mathbf{\in}%
\mathbb{R}
^{512}$ are concatenated and processed \textit{channel-wise,} such that the
concatenated embedding is $\mathbf{w\in}%
\mathbb{R}
^{512\times2}$. $1\times1$ channel-wise convolutions are then applied to
$\mathbf{w}$ and the succeeding activations maps as detailed in Table
\ref{tab:residual_conv}, such that their initial dimensionality ($d=512$) is
retained. As both embeddings $\left\{  \mathbf{\widehat{\varphi}}_{i}%
,\widehat{\mathbf{\phi}}_{i}\right\}  $ are computed by the same
weight-sharing siamese CNN, and are the output of an FC layer, the
corresponding entries in $\mathbf{\widehat{\varphi}}_{i}$ and $\widehat
{\mathbf{\phi}}_{i}$ encode the same \textit{semantic} content. The spatial
locality in such embeddings is uninformative, implying that they can be
processed by either an FC layer or $1\times1$ convolutions, as in the proposed scheme.

The face embeddings $\left\{  \mathbf{\widehat{\varphi}}_{i},\widehat
{\mathbf{\phi}}_{i}\right\}  $ are normalized element-wise before applying the
fusion scheme%
\begin{align}
\mathbf{\widehat{\varphi}}_{i}^{n}  &  =\mathbf{\widehat{\varphi}}_{i}%
.\ast\mathbf{w}_{\varphi}\label{equ:weight}\\
\widehat{\text{ }\mathbf{\phi}}_{i}^{n}  &  =\widehat{\mathbf{\phi}}_{i}%
.\ast\mathbf{w}_{\mathbf{\phi}}\nonumber
\end{align}
where $\left(  \cdot.\ast\cdot\right)  $ denotes pointwise multiplication and
$\mathbf{w}_{\varphi},\mathbf{w}_{\mathbf{\phi}}\mathbf{\in}%
\mathbb{R}
^{512}$. Some of the kinship classes are symmetric in terms of gender (i.e.
B-B, S-S, F-S, M-D), while others are asymmetric (F-D, M-S, SIBS). We utilize
Eq. \ref{equ:weight} to induce asymmetry in the kin verification CNN for the
asymmetric classes, by learning symmetric weights $\mathbf{w}_{\varphi}=$
$\mathbf{w}_{\mathbf{\phi}}=\mathbf{w}$ for the same-gender classes, and
asymmetric weights $\mathbf{w}_{\varphi},$ $\mathbf{w}_{\mathbf{\phi}}$ for
the asymmetric ones.\begin{table}[tbh]
\caption{The proposed embedding fusion subnetwork. The input embeddings
$\mathbf{\widehat{\varphi}}_{i}$ and $\widehat{\mathbf{\phi}}_{i}$, are
channel-wise concatenated. A cascade of $1\times1$ convolutions is applied
along the channels axis. Layers 1D Conv1-1D Conv8 are identical.}%
\label{tab:residual_conv}
\begin{center}
{\small
\begin{tabular}
[c]{l|cccc}\hline
Layer & Kernel & Input & Activation & Output\\\hline\hline
Input &  & $512\times2$ &  & $512\times512$\\%
\begin{tabular}
[c]{l}%
1D Conv1\\
\multicolumn{1}{c}{$\vdots$}\\
1D Conv8
\end{tabular}
& $1\times1$ & $512\times512$ & Relu & $512\times512$\\
1D Conv9 & $1\times1$ & $512\times1,024$ & Relu & $512\times1$\\\hline
\end{tabular}
}
\end{center}
\end{table}

\subsection{Adaptive Sampling of Imbalanced Training Sets}

\label{subsec:adaptive-sampling}

Kinship verification datasets are inherently imbalanced with respect to
multiple attributes. First, families differ significantly by the numbers of
their images and the images per kin \cite{fiwpamiSI2018}. Second, following
Table \ref{table:fiw kin pairs}, the training pairs of some of the classes,
such as F-S, F-D, M-S, and M-D are imbalanced, as most parents (M/F) in the
RFIW dataset have multiple siblings (S/D), and will thus appear in notably
more training pairs than any particular sibling. Thus, uniform sampling of the
training pairs results in oversampling of the parent classes (M/F), and
families having more images than others. Due to the significant imbalances in
the RFIW dataset \cite{fiwpamiSI2018}, applying undersampling or oversampling
is impractical. Downsampling might result in a notably smaller training set,
where kin verification training sets are relatively small to begin with. In
contrast, oversampling might result in overfitting the smaller, oversampled
sets. Let $f_{k}$, $k=1..K$ be the families in the training set, and let
$\left\vert f_{k}\right\vert $ be the number of image pairs of the $k^{\prime
}th$ family. We propose an adaptive sampling scheme, consisting of two steps,
where we first set the \textit{maximal} number of image pairs to be sampled
from each family $\left\vert f_{k}\right\vert _{med}=median\left(  \left\vert
f_{k}\right\vert \right)  $. For families where $\left\vert f_{k}\right\vert
<\left\vert f_{k}\right\vert _{med}$, we only sample $\left\vert
f_{k}\right\vert $ training pairs. To maximize the diversity of the training
set, we choose the training images in the F-D, F-S, M-D, M-S kinship classes
using cyclic buffers, one per family member (F, M, S, D). Thus, for each
training pair, a different pair of images is chosen. \begin{table}[tbh]
\caption{The RFIW dataset \cite{fiwpamiSI2018}. The number of training and
testing kin pairs per kinship class vary significantly.}%
\label{table:fiw kin pairs}%
\centering%
\begin{tabular}
[c]{lccc}\hline
Kinship & Train & Validation & Test\\\hline\hline
Brothers & 29,812 & 55,546 & 18,196\\
Sisters & 19,778 & 35,024 & 4,796\\
Siblings & 28,428 & 15,422 & 9,716\\
Father-Daughter & 41,604 & 35,238 & 15,040\\
Father-Son & 64,826 & 44,870 & 18,166\\
Mother-Daughter & 39,110 & 29,012 & 14,394\\
Mother-Son & 66,464 & 31,094 & 14,806\\
Grfather-Grdaughter & 1,478 & 4,846 & 838\\
Grfather-Grson & 1,388 & 1,926 & 1,588\\
Grmother-Grdaughter & 1,580 & 3,768 & 952\\
Grmother-Grson & 1,284 & 1,844 & 1,470\\
\bottomrule Total & 295,752 & 258,590 & 99,962
\end{tabular}
\end{table}

\section{Experimental Results}

\label{sec:Results}

The proposed scheme was experimentally evaluated using the RFIW (Families In
the Wild) dataset \cite{fiwpamiSI2018}, that was found by Lopez et al.
\cite{7393855} and Damson et al. \cite{Dawson_2019} to be the only unbiased
kin verification dataset. The RFIW is the largest kin verification dataset
depicting $1,000$ different families, for a total of $11,932$ images. There
are $654,304$ positive and negative pairs overall, and the images contain
gender meta-data. We followed the evaluation protocols detailed in
\cite{fiwpamiSI2018}, by applying the \textit{restricted }protocol where the
identities of the subjects in the dataset are unknown, and we are given
predefined pairs of training images, per kinship class. Thus, the adaptive
sampling (Section \ref{subsec:adaptive-sampling}), and the embedding
fine-tuning can't be applied. In the \textit{unrestricted} protocol, both the
identities and the kinship classes (F, M, B, S, etc.) are given, allowing to
train using a significantly larger set of pairs.

Following Robinson et al. \cite{fiwpamiSI2018}, we used a pre-trained VGG-Face
face embedding \cite{VGG_REF} CNN, to compare with previous schemes based on
that backbone. The fusion embedding subnetwork follows Section
\ref{subsec:siamese_asymmetric}. In the \textit{unrestricted} protocol, we
used all of the CNN components detailed in Section \ref{sec:unified-mt}, and
the Sphereface was finetuned using the RFIW dataset. The five-fold
cross-validation split, as in \cite{fiwpamiSI2018}, was used, with no
family-overlap between the different folds. Negative pairs were sampled
randomly, the same number as the positive ones.

\subsection{Network Parameters and Training}

\label{subsec:training}

The proposed approach was trained in two steps, where only the first step
was\ applied in the \textit{unrestricted }protocol. First, the face embedding
CNN was trained using the CASIA-WebFace image dataset \cite{CASIA}. A
SphereLoss was attached to the output of the layer FC-1 and applied using
$l\in\left[  5,1500\right]  $. The learning rate was gradually reduced from
$0.01$ to $0.0001$ in $30$ epochs, using an SGD optimizer.

In the second training step, applied for the \textit{unrestricted }protocol,
the RFIW dataset was used to jointly train a refined face embedding and a
multi-task kin verification classifier. For that, the complete CNN, as
depicted in Fig. \ref{fig:total_arc}, was used. Two SphereLosses were
connected to the outputs of the FC-2 layer, for additional face embedding
refinement, by classifying the faces in the RFIW dataset \cite{fiwpamiSI2018},
and the multi-task kin verification subnetworks were jointly trained.

The MTCNN\ \cite{MTCNN} was used to detect and align the faces in the training
and test images. The face images were scaled to $108\times108$ pixels and
augmented by changing the image gamma factor, scalings $s\in\left[
0.5,1,2\right]  $, horizontal flipping, and random jitter of $dx,dy\in\left[
-2,2\right]  $. All of the augmentations were applied jointly and randomly.
The weighting in Eq. \ref{equ:weight} is applied differently to the symmetric
(B-B, F-S, M-D, S-S) and asymmetric (F-D, M-S, SIBS) pairs of kins. The
training pairs of the symmetric classes are randomly input to either the left
or right side of the CNN. In contrast, each kin in the asymmetric classes is
input to a predefined side of the CNN. We trained the network using the SGD
optimizer with a learning rate $\in$ $\left[  10^{-3},10^{-5}\right]  $ for 30
epochs, where the learning rate was reduced by a factor of 10 when the loss
decay flattened.

\begin{table*}[tbh]
\caption{Kin verification results of the FG2018 challenge \cite{FG2018} using
the \textit{restricted} protocol. The results are given by of the accuracy
percentage.}%
\label{table:challange2018}%
\centering%
\begin{tabular}
[c]{lcccccccccccc}\hline
& M-D & M-S & S-S & B-B & SIBS & GM-GD & GM-GS & F-S & GF-GS & F-D & GF-GD &
Avg.\\\hline\hline
\#1 & 66.6 & 59.84 & 72.9 & 66.2 & 63.0 & 57.6 & 61.1 & 61.5 & 56.92 & 63.3 &
58.0 & 62.4\\
\#2 & 65.8 & 60.46 & 70.2 & 68.4 & 61.6 & 57.0 & 62.8 & 61.2 & 56.4 & 63.1 &
59.42 & 62.4\\
\#3 & 59.5 & 57.81 & 64.7 & 57.2 & 57.4 & 55.4 & 57.7 & 57.4 & 55.7 & 59.4 &
55.9 & 58.0\\
\#4 & 61.5 & 60.5 & 68.0 & 62.7 & 59.0 & 54.5 & 52.2 & 58.6 & 53.3 & 58.21 &
57.9 & 58.7\\
\#5 & 56.7 & 54.1 & 60.2 & 51.3 & 55.4 & 52.4 & 50.2 & 54.1 & 52.5 & 55.9 &
55.1 & 54.4\\
\textbf{ours} & \textbf{73.8} & \textbf{70.9} & \textbf{77.3} & \textbf{69.1}
& \textbf{68.6} & \textbf{63.9} & \textbf{63.3} & \textbf{68.9} &
\textbf{63.7} & \textbf{68.9} & \textbf{61.9} & \textbf{68.2}\\\hline
\end{tabular}
\end{table*}

\subsection{Kin Verification Results Using the RFIW Dataset}

\label{subsec:FIW verification}

The proposed scheme was evaluated by applying it to the RFIW Dataset using
both the \textit{restricted} and \textit{unrestricted} test protocols. We
present the results of the FG2018 challenge \cite{FG2018} using the
\textit{restricted} protocol in Table \ref{table:challange2018}. Our method is
shown to be the most accurate, outperforming the second-best by 13.8\%. In the
\textit{unrestricted} case, we compare to contemporary state-of-the-art
results of the FG2018 \cite{FG2018} challenge. The \textit{unrestricted} test
protocol is more suitable for training deep CNNs, such as ours, as it provides
a larger training set. \begin{table*}[tbh]
\caption{Kinship verification accuracy percentage using the RFIW dataset for
methods applied to the \textit{restricted} and \textit{unrestricted}
protocols. }%
\label{tab:FIW results}
\centering{\small
\begin{tabular}
[c]{lcccccccc}\hline
Method & B-B & S-S & SIBS & F-D & F-S & M-D & M-S & Avg.\\\hline\hline
\multicolumn{9}{c}{Handcrafted features}\\
LBP & 55.5 & 57.5 & 55.4 & 55.1 & 53.8 & 55.7 & 54.7 & 55.4\\
SIFT & 57.9 & 59.3 & 56.9 & 56.4 & 56.2 & 55.1 & 56.5 & 56.9\\\hline
\multicolumn{9}{c}{VGG-Face backbone}\\
+ITML & 57.2 & 61.6 & 57.9 & 58.1 & 54.8 & 57.3 & 59.1 & 57.8\\
+LPP & 67.6 & 66.2 & 71.0 & 62.5 & 61.4 & 65.04 & 63.5 & 65.3\\
+LMNN & 67.1 & 68.3 & 66.99 & 65.7 & 67.1 & 68.1 & 66.2 & 67.0\\
+GmDAE & 68.1 & 68.6 & 67.3 & 66.5 & 68.3 & 68.2 & 66.7 & 67.7\\
+DLML & 68.0 & 68.9 & 68.0 & 66.0 & 68.00 & 68.5 & 67.2 & 67.8\\
+mDML & 69.1 & 70.2 & 68.1 & 67.9 & 66.2 & 70.4 & 67.4 & 68.5\\
\textbf{Ours (VGG)} & \textbf{69.1} & \textbf{77.3} & \textbf{68.6} &
\textbf{68.9} & \textbf{68.9} & \textbf{73.8} & \textbf{70.9} & \textbf{71.1}%
\\\hline
\multicolumn{9}{c}{Face embedding}\\
ResNet-22\cite{RESNET_22_REF} & 66.6 & 69.7 & 60.1 & 59.5 & 60.3 & 61.5 &
59.4 & 62.3\\
VGG-Face\cite{VGG_REF} & 69.7 & 75.4 & 66.5 & 64.3 & 63.9 & 66.4 & 62.8 &
66.7\\
ResNet+CF\cite{RESNET_CF_REF} & 69.9 & 69.5 & 69.5 & 68.2 & 67.7 & 71.1 &
68.6 & 69.2\\
AdvNet\cite{advnet_ref} & 71.8 & 77.4 & 69.8 & 67.8 & 68.8 & 69.9 & 67.3 &
70.4\\
VGG+Tri\cite{vgg_triplet_loss_ref} & 73.0 & 72.5 & 74.4 & 69.4 & 68.2 & 68.4 &
69.4 & 70.7\\
SphereFace\cite{Liu_2017} & 71.9 & 77.3 & 70.2 & 69.3 & 68.5 & 71.8 & 69.5 &
71.2\\\hline
\multicolumn{9}{c}{Face Embedding + Learning}\\
ResNet+SDM\cite{SMDL_ref} & 72.6 & 79.4 & 70.4 & 68.3 & 68.0 & 71.3 & 68.8 &
71.2\\
KinNet\cite{kinnet_ref} & 84.2 & 81.7 & 72.7 & 72.8 & 77.3 & 75.8 & 76.3 &
77.2\\
\textbf{Ours} & \textbf{85.9} & \textbf{86.3} & \textbf{78.0} & \textbf{77.4}
& \textbf{74.9} & \textbf{76.9} & \textbf{75.6} & \textbf{79.6}\\\hline
\end{tabular}
}\end{table*}

The FG2018 results are shown in Table \ref{tab:FIW results}. In the upper part
of the table, we compare with kin verification schemes based on the VGG-Face
\cite{VGG_REF} backbone. We cite the results of methods in which distance
learning was applied on top of the VGG-Face backbone, that were implemented by
Robinson et al. \cite{fiwpamiSI2018}. Our approach was also implemented using
the VGG-Face backbone to allow fair comparison, and outperforms \textit{all}
previous methods, in \textit{all} categories, but for the B-B class. Compared
to the mDML scheme \cite{kinFG2017} having the second-highest average accuracy
among all methods, we achieve an average accuracy improvement of $2.59\%$.

In the mid-segment of Table \ref{tab:FIW results} (\textquotedblleft Face
embedding\textquotedblright), we compare with methods based on face embedding
CNNs. Compared to the Sphereface CNN \cite{fiwpamiSI2018}, our approach shows
an average improvement of $8.34\%$, exemplifying the upside of the proposed
embedding fusion and classification subnetwork. Average improvements of
$10.88\%$ and $6.39\%$ are achieved for the cross and intra-generation
classes, respectively. Our approach is thus more accurate in verifying the
more difficult cross-generation classes. Comparing our scheme implemented
using the VGG-Face \cite{VGG_REF}, to those based on the Sphereface CNN
\cite{fiwpamiSI2018} backbones, it follows that the Sphereface-based ones are
the more accurate. Last, we compared with recent state-of-the-art results
\cite{SMDL_ref,kinnet_ref}, that similar to our method, utilize both face
embedding and metric learning. While these schemes outperform all of the
\textquotedblleft Face embedding\textquotedblright\ methods, they are
outperformed by our approach in all categories.

The results for the FG2020 challenge \cite{robinson2020recognizing} are
reported in Table \ref{table:challange2020} and compared with the leading
entries submitted to the FG2020 challenge, as reported on the challenge
website. We could not apply our approach to the GFGD, GFGS, GMGD, and
GMGS\footnote{GF: grandfather, GM: grandmother, GD: granddaughter, GS:
grandson.} categories, whose training sets are smaller by an order of
magnitude (Table \ref{table:fiw kin pairs}). It follows that our scheme
outperforms all previous results on average, and in five out of eight kinship
categories. \begin{table}[tbh]
\caption{Kin verification results of the FG2020 challenge
\cite{robinson2020recognizing} using the \textit{unrestricted} setup. The
results are given by the accuracy percentage. We compare with the leading
results submitted to the challenge. }%
\label{table:challange2020}
\centering%
\begin{tabular}
[c]{p{0.1in}ccccccccc}\hline
& M-D & M-S & S-S & B-B & SIBS & F-S & F-D & S-S & Avg.\\\hline\hline
\#1 & \textbf{0.78} & 0.74 & 0.8 & 0.8 & 0.77 & 0.81 & 0.75 & 0.8 & 0.77\\
\#2 & 0.75 & 0.74 & 0.77 & 0.77 & 0.75 & 0.81 & 0.74 & 0.77 & 0.76\\
\#3 & 0.75 & \textbf{0.75} & 0.74 & 0.75 & 0.72 & \textbf{0.82} &
\textbf{0.76} & 0.74 & 0.75\\
\#4 & 0.75 & \textbf{0.75} & 0.74 & 0.75 & 0.71 & 0.81 & \textbf{0.76} &
0.74 & 0.75\\
\#5 & 0.74 & \textbf{0.75} & 0.74 & 0.75 & 0.72 & 0.81 & 0.75 & 0.74 & 0.75\\
\textbf{ours} & 0.75 & \textbf{0.75} & \textbf{0.84} & \textbf{0.89} &
\textbf{0.80} & 0.75 & 0.73 & \textbf{0.83} & \textbf{0.78}\\\hline
\end{tabular}
\end{table}

\begin{figure*}[tbh]
\centering%
\begin{tabular}
[c]{cccccccc}
& \textbf{B-B} & \textbf{S-S} & \textbf{F-S} & \textbf{F-D} & \textbf{M-S} &
\textbf{M-D} & \textbf{SIBS}\\
\multicolumn{1}{l}{\textbf{Ref Image}} &
\raisebox{-.5\height}{\includegraphics[width=0.10\textwidth]{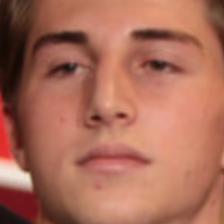}} &
\raisebox{-.5\height}{\includegraphics[width=0.10\textwidth]{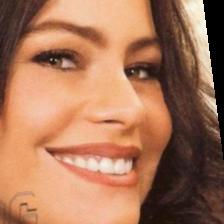}} &
\raisebox{-.5\height}{\includegraphics[width=0.10\textwidth]{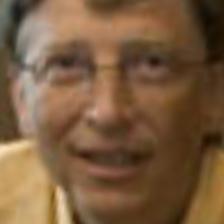}} &
\raisebox{-.5\height}{\includegraphics[width=0.10\textwidth]{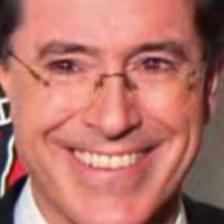}} &
\raisebox{-.5\height}{\includegraphics[width=0.10\textwidth]{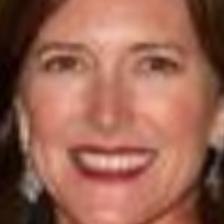}} &
\raisebox{-.5\height}{\includegraphics[width=0.10\textwidth]{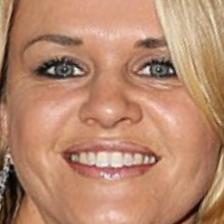}} &
\raisebox{-.5\height}{\includegraphics[width=0.10\textwidth]{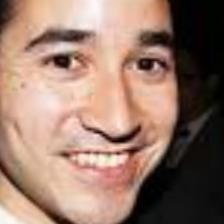}}\\
\multicolumn{1}{l}{\textbf{True Positive}} &
\raisebox{-.5\height}{\includegraphics[width=0.10\textwidth]{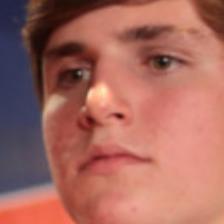}} &
\raisebox{-.5\height}{\includegraphics[width=0.10\textwidth]{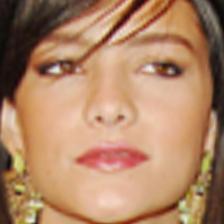}} &
\raisebox{-.5\height}{\includegraphics[width=0.10\textwidth]{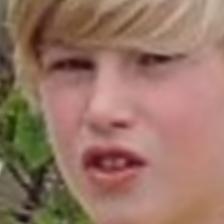}} &
\raisebox{-.5\height}{\includegraphics[width=0.10\textwidth]{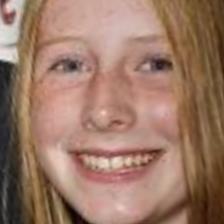}} &
\raisebox{-.5\height}{\includegraphics[width=0.10\textwidth]{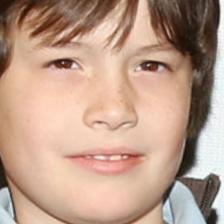}} &
\raisebox{-.5\height}{\includegraphics[width=0.10\textwidth]{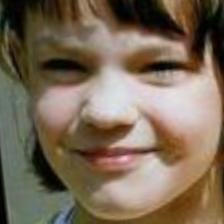}} &
\raisebox{-.5\height}{\includegraphics[width=0.10\textwidth]{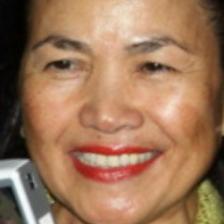}}\\
\multicolumn{1}{l}{\textbf{False Negative}} &
\raisebox{-.5\height}{\includegraphics[width=0.10\textwidth]{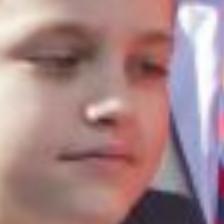}} &
\raisebox{-.5\height}{\includegraphics[width=0.10\textwidth]{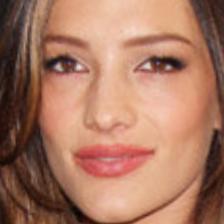}} &
\raisebox{-.5\height}{\includegraphics[width=0.10\textwidth]{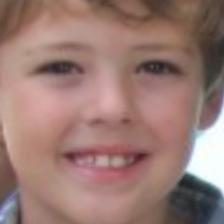}} &
\raisebox{-.5\height}{\includegraphics[width=0.10\textwidth]{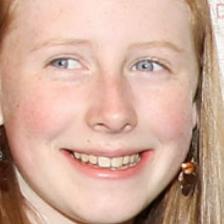}} &
\raisebox{-.5\height}{\includegraphics[width=0.10\textwidth]{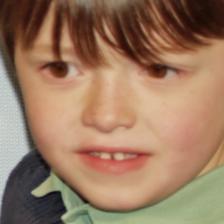}} &
\raisebox{-.5\height}{\includegraphics[width=0.10\textwidth]{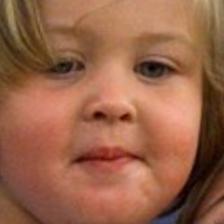}} &
\raisebox{-.5\height}{\includegraphics[width=0.10\textwidth]{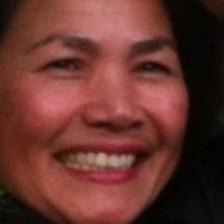}}\\
\multicolumn{1}{l}{\textbf{True Negative}} &
\raisebox{-.5\height}{\includegraphics[width=0.10\textwidth]{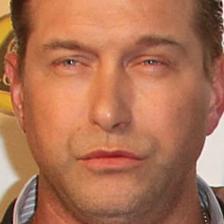}} &
\raisebox{-.5\height}{\includegraphics[width=0.10\textwidth]{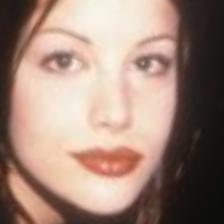}} &
\raisebox{-.5\height}{\includegraphics[width=0.10\textwidth]{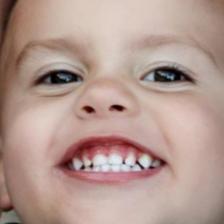}} &
\raisebox{-.5\height}{\includegraphics[width=0.10\textwidth]{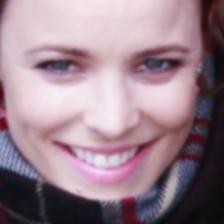}} &
\raisebox{-.5\height}{\includegraphics[width=0.10\textwidth]{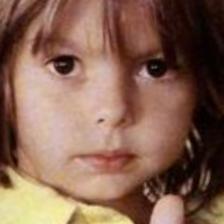}} &
\raisebox{-.5\height}{\includegraphics[width=0.10\textwidth]{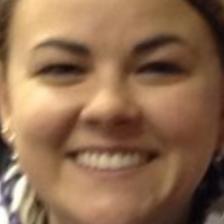}} &
\raisebox{-.5\height}{\includegraphics[width=0.10\textwidth]{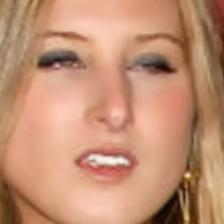}}\\
\multicolumn{1}{l}{\textbf{False Positive}} &
\raisebox{-.5\height}{\includegraphics[width=0.10\textwidth]{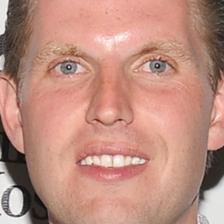}} &
\raisebox{-.5\height}{\includegraphics[width=0.10\textwidth]{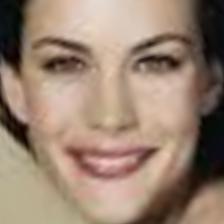}} &
\raisebox{-.5\height}{\includegraphics[width=0.10\textwidth]{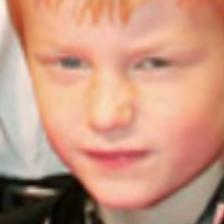}} &
\raisebox{-.5\height}{\includegraphics[width=0.10\textwidth]{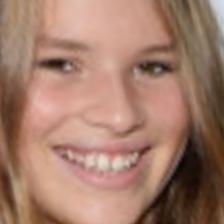}} &
\raisebox{-.5\height}{\includegraphics[width=0.10\textwidth]{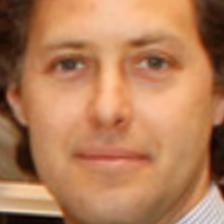}} &
\raisebox{-.5\height}{\includegraphics[width=0.10\textwidth]{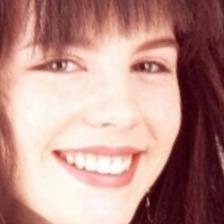}} &
\raisebox{-.5\height}{\includegraphics[width=0.10\textwidth]{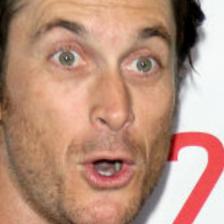}}
\end{tabular}
\caption{Kin verification examples from the RFIW dataset \cite{fiwpamiSI2018},
as classified by the proposed scheme. The upper row shows a face image, and
the succeeding rows depict the corresponding kin verification results.}%
\label{fig:examples1}%
\end{figure*}

Qualitative kin verification examples of images from the RFIW dataset are
depicted in Fig. \ref{fig:examples1}. The reference images were chosen
randomly and depict the common results. For the S-S and B-B kinship classes,
the face similarity might seem intuitive. But the similarity of other kinship
classes such as F-S, F-D, etc., seems to be unintuitive, In particular,
comparing the True Positives with the False Positives classifications does not
provide insights, in contrast to face verification results.

\subsection{Ablation Study}

\label{subsec:Ablation}

We conducted an ablation study reported in Table \ref{table:abalation},
wherein each experiment, we modified a single algorithmic component. Our
approach was applied following the \textit{unrestricted} protocol using the
RFIW dataset, as in Section \ref{subsec:FIW verification}.

\textbf{Adaptive sampling}. We first studied the adaptive sampling scheme
presented in Section \ref{subsec:adaptive-sampling}. The statistics of the
RFIW dataset \cite{fiwpamiSI2018}, with and without applying the adaptive
sampling, are reported in Table \ref{tab:FIW dataset} and exemplify the
balancing effect. The variations in the mean and standard deviation of the
balanced RFIW are notably smaller. To compare the corresponding kin
verification accuracy, we trained a network following our approach, without
the adaptive sampling. Instead, we used the common random sampling. The
proposed adaptive sampling is shown to improve the classification accuracy by
a significant $4.44\%$. \begin{table}[tbh]
\caption{The number of images per family and family member in the RFIW dataset
\cite{fiwpamiSI2018}, with and without applying the adaptive sampling
introduced in Section \ref{subsec:adaptive-sampling}. Using the adaptive
sampling, the dataset's statistics are balanced.}%
\label{tab:FIW dataset}%
\centering%
\begin{tabular}
[c]{c|ccc|ccc}\hline
& \multicolumn{3}{c|}{\#Images per family} & \multicolumn{3}{c}{\#Images per
family member}\\
Fold\# & \ Maximum & Mean & Std & Maximum & Mean & Std\\\hline\hline
& \multicolumn{6}{c}{\textbf{RFIW}}\\
$1$ & 1,312 & 111 & 160 & 656 & 34 & 56\\
$2$ & 1,532 & 114 & 204 & 722 & 36 & 66\\
$3$ & 760 & 97 & 136 & 380 & 30 & 43\\
$4$ & 15,132 & 265 & 1,571 & 2,688 & 76 & 297\\
$5$ & 6,808 & 190 & 699 & 2,373 & 57 & 169\\\hline
& \multicolumn{6}{c}{\textbf{Balanced RFIW}}\\
$1$ & 1,600 & 125 & 224 & 387 & 37 & 52\\
$2$ & 1,600 & 121 & 213 & 540 & 36 & 54\\
$3$ & 1,600 & 118 & 199 & 380 & 36 & 51\\
$4$ & 1,532 & 116 & 193 & 722 & 35 & 60\\
$5$ & 1,312 & 112 & 174 & 656 & 35 & 59\\\hline
\end{tabular}
\end{table}

\textbf{Multi-task training}. The multi-task training was evaluated by
training a different CNN for each kinship class instead of the unified
approach, as in Section \ref{sec:unified-mt}. Each of the multiple single-task
face-embedding CNNs was refined as in the first phase in Section
\ref{subsec:training}. For that, we used the RFIW dataset and enabled one of
the Sphere losses ${L}_{Sphere}^{\mathbf{\widehat{\varphi}}_{i},\widehat
{\mathbf{\phi}}}$, as in Fig. \ref{fig:total_arc}. The fusion and
classification subnetworks were then trained separately per kinship class. The
results show that the proposed multi-task training provides an average
accuracy improvement of $1.1\%$, implying that different kinship classes share
a joint structure utilized by the proposed scheme to improve accuracy.

\textbf{Embedding-fusion scheme}. The embedding-fusion subnetwork, introduced
in Section \ \ref{subsec:siamese_asymmetric}, was studied by replacing it with
the common fusion approach where the embeddings $\mathbf{\widehat{\varphi}%
}_{i},\widehat{\mathbf{\phi}}\in%
\mathbb{R}
^{512}$ were concatenated, followed by an FC layer. This results in a
significant accuracy degradation of close to 25\%, compared with our scheme.
We attribute that to overfitting, as kin verification is characterized by
significant intraclass variability. We also applied our approach without using
a fusion scheme, such that the face embeddings were used for kin verification,
as in \cite{fiwpamiSI2018}. In such a scheme, there is no overfit, but the
results are inferior by $5.9\%$.

\textbf{Asymmetric normalization}. We studied the use of the asymmetric
normalization in Eq. \ref{equ:weight} by applying the proposed scheme without
the normalization, and by using symmetric weights $\mathbf{w}_{\varphi
}=\mathbf{w}_{\varphi}$. Using the asymmetric weights improves the average
accuracy by $0.65\%$ and $0.57\%,$ compared to not applying the weighting, and
using symmetric weights, respectively. \begin{table*}[tbh]
\caption{Ablation study results. The results are given by of the accuracy
percentage.}%
\centering%
\begin{tabular}
[c]{l|cccccccc}\hline
& B-B & S-S & SIBS & F-S & F-D & M-S & M-D & Average\\\hline\hline
Fusion by concatenation & 57.18 & 55.01 & 53.71 & 57.97 & 54.23 & 55.72 &
55.74 & 55.65\\
Face embedding only & 77.35 & 81.22 & 70.19 & 69.86 & 72.67 & 71.77 & 72.50 &
73.65\\
FIW\ random sampling & 78.33 & 84.35 & 70.23 & 71.64 & 74.43 & 73.06 & 73.00 &
75.00\\
Single task training & 84.99 & 84.96 & 74.63 & 75.07 & 74.64 & 76.74 & 77.96 &
78.42\\
No normalization & 85.69 & 86.03 & 74.56 & 76.36 & 75.29 & 76.14 & 78.25 &
78.90\\
Symmetric normalization & 85.77 & 86.14 & 74.50 & 77.25 & 75.34 & 75.87 &
78.05 & 78.98\\
\textbf{Ours} & \textbf{85.89} & \textbf{86.26} & \textbf{74.86} &
\textbf{77.39} & \textbf{75.59} & \textbf{76.89} & \textbf{79.98} &
\textbf{79.55}\\\hline
\end{tabular}
\label{table:abalation}%
\end{table*}

\section{Conclusions}

\label{CONCLUSION}

In this work, we proposed a multi-task deep learning-based approach for kin
verification, where all kinship classes are jointly trained, utilizing all of
the training data. We also introduced a novel embedding fusion scheme to fuse
the face embeddings of the input subjects, that is shown to avoid the
overfitting issues common in kin verification, and provides notable accuracy
improvement. An adaptive sampling of the training pairs in the RFIW dataset
allows creating a balanced training set that further improves the kin
verification accuracy. Our scheme is experimentally shown to achieve
state-of-the-art accuracy when applied to the RFIW, FG2018, and FG2020
datasets in both the \textit{restricted} and \textit{unrestricted} test protocols.

\bibliographystyle{IEEEtran}
\bibliography{kinface_short_plain}

\end{document}